\documentclass[sigconf]{acmart}

\AtBeginDocument{%
  \providecommand\BibTeX{{%
    \normalfont B\kern-0.5em{\scshape i\kern-0.25em b}\kern-0.8em\TeX}}}

\usepackage{bm}
\usepackage{booktabs}
\usepackage{graphicx}
\usepackage{epsfig}
\usepackage{algorithm, algpseudocode}
\usepackage{multirow}
\usepackage{makecell}
\usepackage{color}
\usepackage{subfigure}
\usepackage{mathrsfs}
\usepackage{amsmath}
\usepackage{url}

\newenvironment{pequation}{\small\begin{eqnarray*}}{\end{eqnarray*}}

\setcopyright{acmlicensed}

\copyrightyear{2020} 
\acmYear{2020} 
\setcopyright{acmlicensed}\acmConference[KDD '20]{Proceedings of the 26th ACM SIGKDD Conference on Knowledge Discovery and Data Mining}{August 23--27, 2020}{Virtual Event, CA, USA}
\acmBooktitle{Proceedings of the 26th ACM SIGKDD Conference on Knowledge Discovery and Data Mining (KDD '20), August 23--27, 2020, Virtual Event, CA, USA}
\acmPrice{15.00}
\acmDOI{10.1145/3394486.3403355}
\acmISBN{978-1-4503-7998-4/20/08}
\begin{document}

\fancyhead{}

\title{Balanced Order Batching with Task-Oriented Graph Clustering}

\author{
	Lu Duan{$^{1*}$},
	Haoyuan Hu{$^{1*}$}, 
	Zili Wu{$^1$},
	Guozheng Li{$^1$},
	Xinhang Zhang{$^1$},
	Yu Gong{$^2$}, 
	Yinghui Xu{$^1$}
}
\thanks{\large{$^*$} \footnotesize{Lu Duan and Haoyuan Hu contribute equally, and  Haoyuan Hu is the corresponding author}}
\affiliation{
	\institution{$^{1}$Zhejiang Cainiao Supply Chain Management Co. Ltd \quad $^{2}$Alibaba Group}
}

\email{{duanlu.dl,zili.ziliwu,guozheng.lgz,
		xinhang.zxh,haoyuan.huhy,renji.xyh}@cainiao.com, gongyu.gy@alibaba-inc.com}
\begin{abstract}
	 Balanced order batching problem (BOBP) arises from the process of warehouse picking in Cainiao, the largest logistics platform in China. Batching orders together in the picking process to form a single picking route, reduces travel distance. The reason for its importance is that order picking is a labor intensive process and, by using good batching methods, substantial savings can be obtained. The BOBP is a NP-hard combinational optimization problem and designing a good problem-specific heuristic under the quasi-real-time system response requirement is non-trivial. In this paper, rather than designing heuristics, we propose an end-to-end learning and optimization framework named Balanced Task-orientated Graph Clustering Network (BTOGCN) to solve the BOBP by reducing it to balanced graph clustering optimization problem. In BTOGCN, a task-oriented estimator network is introduced to guide the type-aware heterogeneous graph clustering networks to find a better clustering result related to the BOBP objective. Through comprehensive experiments on single-graph and multi-graphs, we show: 1) our balanced task-oriented graph clustering network can directly utilize the guidance of target signal and outperforms the two-stage deep embedding and deep clustering method; 2) our method obtains an average $4.57$m and $0.13$m picking distance\footnote{"m" is the abbreviation of the meter (the SI base unit of length) .} reduction than the expert-designed algorithm on single and multi-graph set and has a good generalization ability to apply in practical scenario.
\end{abstract}






\maketitle

\section{Introduction}
\label{sec:intro}
\textit{Order picking} is a critical component of warehouse operations as shown in Figure 1, which affects customer service, logistic costs and even the efficiency of whole supply chain. It is the most labor-cost-intensive operation which takes up 50\%- 70\% \cite{frazelle2002world,tompkins2010facilities} of the overall warehouse operating cost. In our case, Cainiao, the largest logistics platform in China, which on average deals with a hundred million logistics orders in warehouses each day, is very concerned about improving the warehouse operating efficiency, especially the order picking part, since 1\% improvement of it will reduce millions of dollars cost. 

Due to its higher efficiency compared to \textit{discrete order picking} (\textit{pick-by-order}), \textit{batch picking} have been become the most prevalent mode of manual order picking, where items of several orders in one batch can be collected simultaneously on a single tour. The process of grouping a set of orders into pick lists (batches) is referred to as \textit{order batching} and correspondingly, the optimization problem researching how to divide orders into different batches to obtain the shortest total picking distance is referred to as \textit{order batching problem} (OBP). OBP is known to be NP-hard in the strong sense\cite{gademann2005order}. Actually, OBP has some similar features with the well-known combinational optimization problem, i.e., capacitated vehicle routing problem (CVRP) \cite{ralphs2003capacitated}, but differs from that with respect to the order's impartibility, i.e. items in one order must be picked up in one batch, which makes the OBP more complex than CVRP. Besides, in our case, the number of orders in each batch must be equal, for the picking device always load a fixed number of picking baskets, one basket for one order, in practice. Therefore, we focus on the \textit{balanced order batching problem} (BOBP) in this paper.
\begin{figure}[th]
	\centering
	\includegraphics[angle=0, width=0.82\columnwidth]{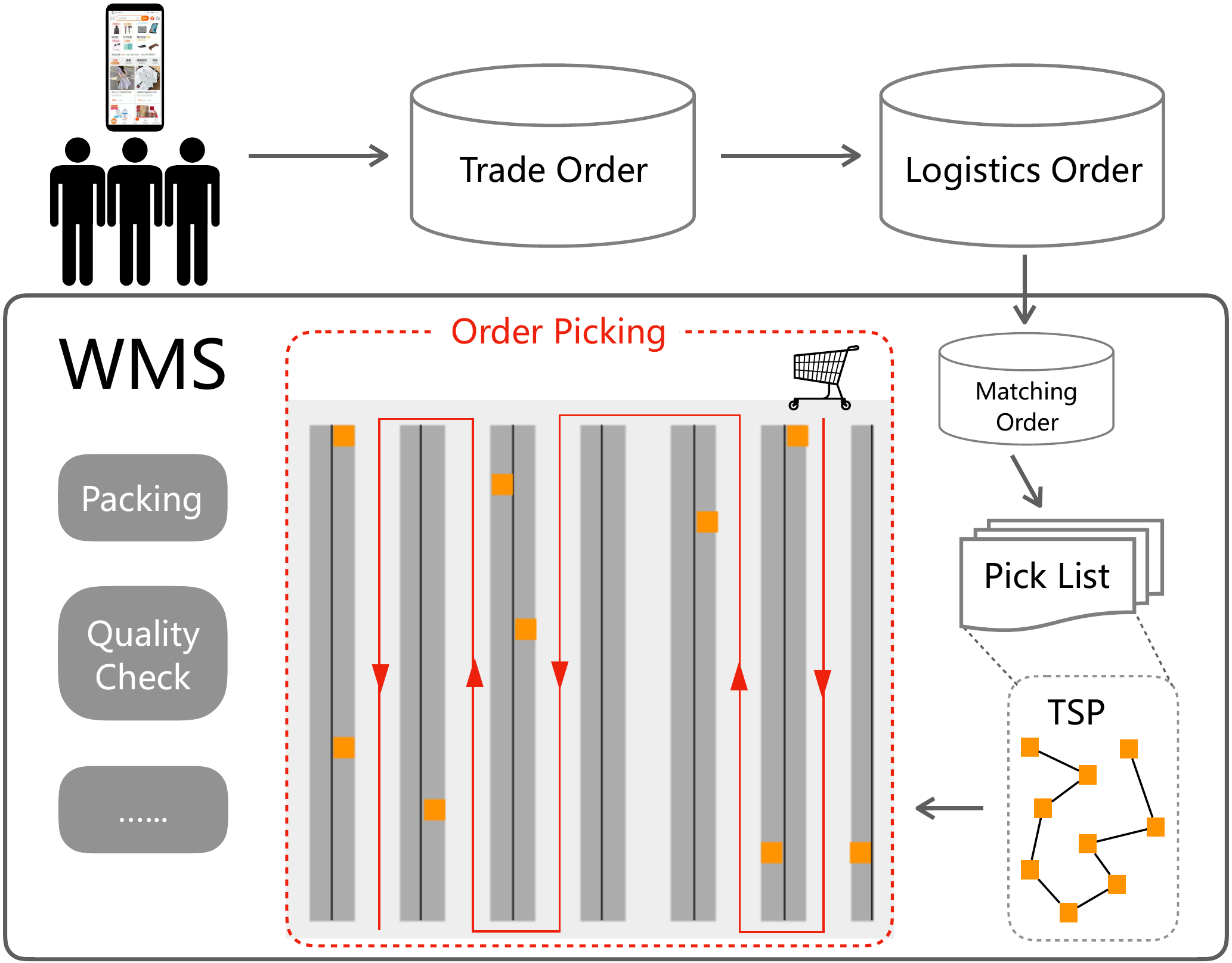}
	\caption{The work-flow of the order processing. The generation of pick list will go through a process from coarse to fine . Coarse: 500 orders are filtered from the order pool. Fine: the filtered orders can be combined (batched) into to pick lists (batches) until the capacity of the device is exhausted and the  performance indicator of pick list is the distance of picking route generated by a TSP algorithm. 
	In our case, the device capacity is set up as a fixed number of orders. After pick lists generation, the logistic orders will go through picking, packing, quality check before delivery. All these operations are included in our Cainiao warehouse management system (WMS). }
	\label{fig:sys}
\end{figure}

In operation research community, OBP has been extensively studied for years. In general, there are two main classes of approaches: exact algorithms based on mathematical modeling and  human-designed heuristics methods. Although exact algorithms can guarantee the optimality, they can only solve very limited scale problems (\cite{henn2010metaheuristics} reports only 50 orders in 900 storage locations) without applicability in real-world. So, many heuristics methods are elaborately designed by experts with in-depth domain knowledge for practical application. However, with the application of our IoT device LEMO \footnote{https://iot.cainiao.com/portal/lemo}, a sort of RF scanner invented by Cainiao for picking orders without papery pick list, heuristics methods can not provide good enough solution for such vast amount of orders under the quasi-real-time system response requirement. Because constructive heuristics cannot guarantee the solution quality and meta-heuristics need many complex algorithm iterations to get a competitive solution.

To overcome above weakness, we try to use the machine learning methods to solve combinational optimization problem in view of its fast online inference ability. However, the pure learning methods which directly predict a optimization problem solution using supervised learning or reinforcement learning (RL) \cite{bello2016neural,kool2018attention,vinyals2015pointer} have been proven to have difficult in discovering the algorithmic structure in the huge solution space even provided with large scale training data.  Accordingly, the majority of solutions  \cite{bahulkar2018community,berlusconi2016link,burgess2016link,tan2016efficient,yan2012detecting}adopt the two-stage approaches which train machine learning model firstly and then input the model predications to an optimization algorithm to get the final task result. However, in usual, two-stage approaches only get a sub-optimal solution. Therefore, some end-to-end approaches \cite{wilder2019melding,wilder2019end,gao2019deep} have been developed recently, where the differentiable optimization algorithm layer is integrated into the learning architecture so as to combine the merits of learning and optimization methods. However, most of these works focus on the universal framework with general learning methods, such as auto-encoder (AE) \cite{gao2019deep} or primal graph convolutional networks (GCN) \cite{wilder2019end}. Obviously, due to the explicit hierarchical structure and multiple relations between items and orders in BOBP, we need a learning method with more strong expression ability. Furthermore, these works commonly use a continuous relaxation of the discrete problem to propagate gradients through the optimization procedure, while we prefer a more direct and exact method to handle the non-differentiable problem. Therefore, we introduce an improved learning and optimization end-to-end framework named \textit{Balanced Task-oriented Graph Clustering Network} (BTOGCN) to solve the BOBP.

Since the BOBP naturally can be regarded as a variant of clustering problem, we construct a heterogeneous order-item (order batching) graph 
so that the BOBP is transformed into a special graph clustering task on this graph with cluster balance constraint and the non-differentiable task loss. For better discovering the structure information hidden in the hierarchal heterogeneous graph of BOBP, we devise a type-aware heterogeneous graph convolutional networks (HetGNN). Then the embeddings of orders learned by HetGNN are clustered in balance by a differentiable optimization layer. However, the task-based loss (total picking distance) which is the objective of BOBP in our case, cannot be derived with respect to the cluster results (order batches) because of the optimization process used to calculate the loss is related to the item set of clusters. As a result, we propose a task-oriented estimator network to approximate the task-based loss, which makes it possible to train our model in end-to-end way.

In summary, we list the contributions of this work as below:

\begin{itemize}
	\item First and foremost, to our best knowledge, we are the first one to propose a task-oriented end-to-end graph learning and optimization framework to solve the BOBP. In the meanwhile, our approach has shown its potential to solve many classical combinational optimization problems, such as VRP and bin packing problem.
	\item Next, we use the objective of BOBP as the task-loss to guide the graph clustering. However, the task-loss is non-differentiable with respect to batches for the sake of the set union operation and picker route optimization process. In order to tackle this problem, we propose a task-oriented estimator network to approximate the task-based loss, which enables our model being trained in the unsupervised manner.
	\item Then, in view of the natural hierarchical structure of BOBP, we design a type-aware heterogeneous graph convolutional networks (HetGNN) on balanced order batching graph with different types of nodes and edges. And the HetGNN has been proven its superior representation learning ability combined with a simple k-means layer and greedy assignment to obtain better solutions.
	\item 
	As a result, we have saved 4.57m and 0.13m average picking distances than expert-designed heuristic algorithm \cite{ho2008order} on single graph and multi-graph dataset, respectively. Numerical results also demonstrate that our BTOGCN method significantly outperforms the two-stage models without task goal and has a good generalization ability to apply in practical scenario.
\end{itemize}


\section{Related Work}
\label{sec:related}

Motivated by recent studies on decision-focused learning\cite{donti2017task,wilder2019melding,wilder2019end,gao2019deep} which include optimization process into training architecture to improve performance of downstream decision, we propose an end-to-end learning and optimization framework BTOGCN to solve a very complex combinational optimization problem named BOBP. However, our task has some complex and problem-specific challenges to tackle compared to the aforementioned universal task.

First, our task loss cannot be directly calculated based on the decision solution, because in BOBP, the computation of picking distance with respect to the batch needs extra algorithm, specifically, we have to union the items of orders in one batch at first, then solve a traveling salesman problem (TSP) to get the picking distance for these items. Obviously, the set union operation and TSP algorithm make the task loss non-differentiable with respect to the discrete batch. On the contrary, in \cite{wilder2019end} which combines the representation learning with explicit soft k-means algorithm, the loss function can not only be directly computed in the forward pass, but also naturally differentiable to the k-means assignment matrix $r$, because the original discrete cluster is continuously relaxed in their model all the time. The similar relaxation approach is also applied in \cite{wilder2019melding} to simplify the problem. Thus, we must calculate the task loss in a faster way to improve the training efficiency and solve the derivative problem in other way. In fact, similar problems are often countered in RL, thus the Q-network is devised to predict the future cumulative reward in a sequential decision-making process given current state and action\cite{lillicrap2015continuous}. With the growing concern with the decision-based learning, \cite{chen2019task} proposes an end-to-end learning scheme which integrates the non-differential decision evaluation process into the learning architecture via a task-goal estimator. We follow the idea of task-based estimator in our work.
 
Second, in view of the similarity of BOBP with clustering problem, we take into consideration of using deep clustering methods to get batches. In recent years, numerous deep clustering methods  \cite{xie2016unsupervised,yang2017towards,gao2019deep,genevay2019differentiable} are proposed, which aim to find a "cluster-friendly latent embedding space by deep learning to make a better cluster decision in an end-to-end way. To cluster the data, these works usually will directly learn the cluster centroid, because the cluster centroids is relatively stable in embedding space as a result of the changeless data distribution, so deep clustering is often applied in community detection and image classification \cite{cavallari2017learning,perez2017effectiveness}. However, in our scene, the cluster centroids might dramatically change according to the orders waiting to be picked. Therefore, an explicit and non-parameter clustering algorithm layer is what we need, such as k-means. But as we all know, in k-means, the derivative of cluster assignment with respect to data cannot be directly computed. Thus, \cite{gao2019deep} adapts the Gumbel-Softmax re-parameterization trick to estimate the gradient, \cite{genevay2019differentiable} transforms the clustering problem to optimal transport(OT) problem  and \cite{wilder2019end} transforms the cluster centroid update formula of k-means to a helper function and use the implicit function theorem to get that derivation. Nevertheless, these methods cannot provide a balanced discrete cluster solution for us. Therefore, we add a dynamic assignment layer in the behind of the differential k-means layer to keep the balance of clusters.

Third, we need a more powerful representation learning technique to mine intricate relations among orders and items than AE, which is commonly seen in varieties of end-to-end pipelines \cite{xie2016unsupervised,guo2017improved,yang2017towards}. 
GCN has proved its power in many real-world application such as recommendation system\cite{wang2018billion} and chemical properties prediction\cite{shang2018edge}. However, most of successful GCN works\cite{hamilton2017inductive,velivckovic2017graph} focus on the homogeneous graph, while data represented as heterogeneous graph is more natural in many cases. To the best of our knowledge, some new works \cite{shang2018edge,liu2018heterogeneous} has applied the heterogeneous graph to get success. \cite{shang2018edge} proposed a model named EAGCN which applies the "multi-attention" mechanism to process the heterogeneous graph where each attention only aggregate neighbors connected by the edges of corresponding type. 
Similarly, \cite{liu2018heterogeneous} exploits the heterogeneous graph with attention mechanism, but the graph contains multiple types of nodes rather than edges. 
However, in our BOBP, we need to design a graph embedding architecture by combining information of different types of nodes and edges.

As detailed above, these three key challenges brought by the complexity of BOBP are elaborately taken care in our BTOGCN approah, to the best of our knowledge, there is no one end-to-end framework with the ability to solve the BOBP solely yet.

\section{Problem Statement}
\label{sec:problem}

In this section, we first give an informal description about how the BOBP works in our warehouse management system (WMS). Then we will give a formal definition and optimization objective for BOBP, and discuss how to transform it into balanced graph clustering optimization problem. 

\subsection{Balanced Order Batching Problem}


In Cainiao, an average of 100 million express orders pass through the warehouses every day. Figure \ref{fig:sys} illustrates the work-flow of order processing. After the customer places an order on the e-commerce website, e.g. TaoBao or TMall, it will be converted to a logistics order (consisting of several items) to be processed in the warehouse. For the sake of simplification, the order specifies logistics order afterwards. At each decision-making time, there are tens of thousands of logistic orders waiting to be processed in the order pool. As a well-known NP-hard problem \cite{gademann2005order}, it is very expensive to generate pick lists from such a huge candidate set. In order to reduce the computational complexity and meet the requirements of time efficiency, in practice, we choose the coarse-to-fine framework commonly seen in the traditional recommendation system scenario. Firstly, a coarse model (deep matching model) will be deployed to generate hundreds of orders (usually 500) from the current order pool, then the generated candidate set will be processed by a fine model (order batching network) to produce pick lists. In our paper, we concentrate on the process of grouping the orders in candidate set into batches on candidate set (matching pool) after matching procedure. Essentially, the order batching network consists of the following two components:
\begin{itemize}
	\item order clustering, i.e. the transformation of orders into pick lists, which cannot exceed the capacity of the device, here the capacity is set to a fixed number of orders;
	\item picker routing, i.e. planning the corresponding picking route for the items in each pick list by solving it as TSP.
\end{itemize}

Once clusters of orders have been formed, the calculation of the travel distance for the routes requires a number of TSP solutions (one route for one batch). The balanced order batching problem in a multiple block warehouse that we analyze in this paper is reduced to the splitting of the $N$ orders into $K$ batches, each with $c$ orders, so as to minimize the total picking distance. Hence, we are implicitly assuming that $N=c\times K$, which is referred as a balanced order batching problem. 


\subsection{Balanced Graph Clustering Optimization Perspective}

Given a set of candidate $N$ orders $O = \{o_1,\dots,o_N\}$, our goal is to divide them into $K$ batches $S=\{s_k,1 \leq k \leq K\} \subseteq B$ ($B$ is the set of all feasible batches), so that the total travel distance of all batches denoted as  $\bm{d_S}=\sum_{k=1}^{K}d_k$ is minimized. In particular, the $d_k$ is the distance calculated by solving the problem of finding a shortest route to retrieve all items of the batch $s_k$ as a traveling salesman problem. Each feasible batch $b_m \in B$ is characterized by a zero-one vector $a_m = (a_{1m},\dots,a_{Nm})$, where $a_{jm} = 1$ if order $o_j$ is included in batch $b_m$ $(b_m\in B, o_j\in O)$. Besides, binary decision variable $x_{m}$ is set to 1 if batch $b_m$ is chosen to $S$, or $x_m=0$, otherwise. Note that, in our case, picking device capacity is expressed as a fixed number $c$ of orders in a batch, so that batch $b_m$ is feasible only if $\sum_{o_j\in O}a_{jm} = c$. Based on the above descriptions of problem and notations, the mathematical model of the BODP is formulated as follows:
\begin{pequation}
	\min\,\,  \sum_{b_m\in B}d_mx_m
\end{pequation}
subject to:
\begin{scriptsize}
	\begin{align} 
	\sum_{b_m\in B}a_{jm}x_m&=1, \, \, \forall{o_j\in O}; \\
	\sum_{b_m\in B}x_m&=K; \\
	\sum_{o_j\in O}a_{jm}&=c, \, \, \forall{b_m \in B}; \\
	x_m\in\{&0,1\}, \, \, \forall{b_m \in B}.
	\end{align}
	\vspace{-10pt}
\end{scriptsize}

Equation (1) ensures that each order is exactly assigned to one chosen batch which usually referred as integrity condition. Equation (2) and Equation (3) ensure that $K$ batches are chosen and each of them has $c$ orders.

\begin{figure}[th]
	\vspace{-10pt}
	\centering
	\includegraphics[angle=0, width=0.84\columnwidth]{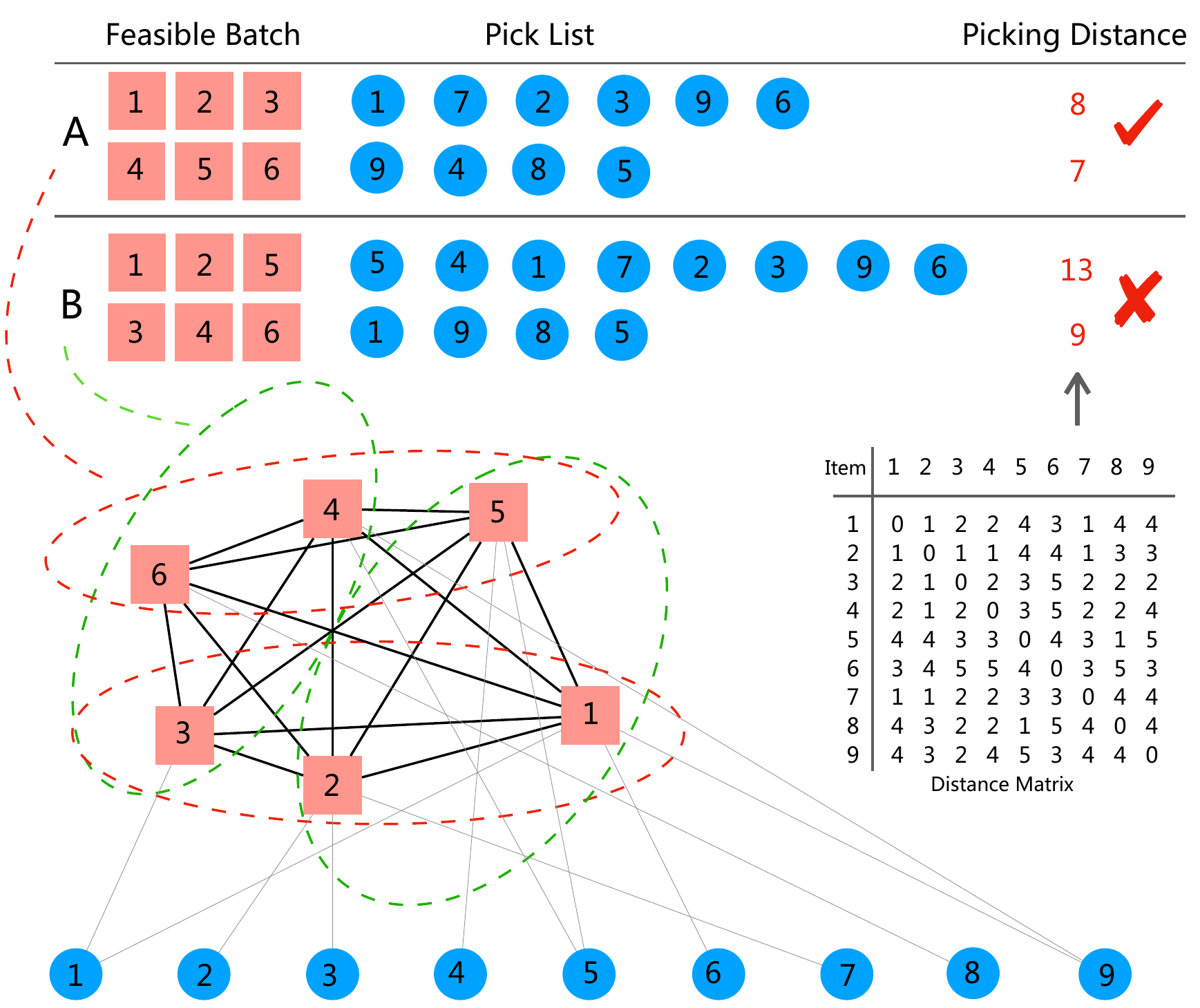}
	\caption{Illustration for a specific heterogeneous graph $\mathbb{G}$ with $N = 6$ and $K = 2$ and $c = 3$. The orange squares  and  blue circles represent orders and items, respectively. The black lines and grey dashed lines represent order-to-order and order-to-item edges. For the sake of beauty, all the item-to-item edges are not drawn. We show two kinds of clustering solutions in graph (red and green) and the corresponding feasible batches (A and B). We suppose that $\bm{d_B} > \bm{d_A}$ means that A clustering is better than B given candidate order set $O$ and item set $I$.}
	\label{het}
	\vspace{-10pt}
\end{figure}
A heterogeneous graph $\mathbb{G}(\mathcal{V}, \mathcal{E})$ consists of a set of vertices $\mathcal{V}$ and a set of edges $\mathcal{E}$. There is a set of node types $A$, and each vertex $v \in \mathcal{V}$ belongs to one of the node types, denoted by $\phi(v) = p \in A$, where $\phi$ is the mapping function from $\mathcal{V}$ to $A$. We represent an edge $e \in \mathcal{E}$ from the vertex $i \in V$ to $j \in \mathcal{V}$ with a relation type $r$ as a triplet $e = (i, j, r)$, where $r \in R$ and $R$ is the set of relation types. 
For a vertex $j$ and edge type $r$ , the set of linkages with its neighboring nodes is defined as $\mathcal{E}_{j,r} =\{(i,j,r') \in \mathcal{E}|i\in V,r' = r\}$. Correspondingly, we can conduct a heterogeneous order batching graph $\mathbb{G}$  with $\mathcal{V}=\{\mathcal{O},\mathcal{I}\}$, where $\mathcal{O}$ is the set of candidate orders, $\mathcal{I}$ is the set of item nodes included in the candidate orders and then we simply define 
three types of relation $R=\{oo,oi,ii\}$, i.e. order-to-order, order-to-item and item-to-item. The detailed edge description is as follows: 
\begin{itemize}
	\item order-to-order edge, i.e. the picking route consisting of items set in two different orders;
	\item order-to-item edge, i.e. indicating whether the order $o$ contains the item $i$;
	\item item-to-item edge, i.e. the distance between two different items.
\end{itemize}

You can take Figure \ref{het} as an example. In conclusion, we can transform the balanced order batching problem into balanced graph clustering optimization problem. That is to say,
we aim to find a best way to divide the order nodes $\mathcal{O}$ in $\mathbb{G}$ into $K$ clusters(batches) so that the corresponding total picking distance calculated by items retrieving route is minimized.

\section{Approach: Balanced Task-oriented Graph Clustering Network}
\label{sec:model}
\begin{figure*}[h]
	\centering
	\includegraphics[angle=0, width=1.9\columnwidth]{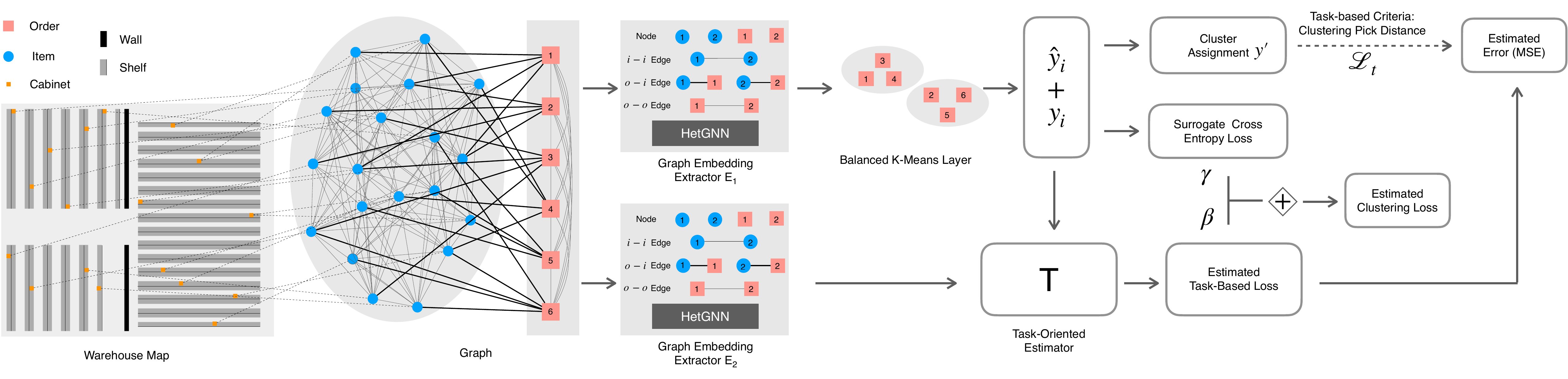}
	\caption{Overview of the Balanced Task-oriented Graph Clustering Network.}
	\label{fig:model}
\end{figure*}
In Section 3.2, we firstly introduce our special constructed heterogeneous graph $\mathbb{G}(\mathcal{V}, \mathcal{E})$ which consists of 2 node types (order, item) and 3 edge types ($oo, oi, ii$), and then convert BOBP into a problem with regard to how to divide the $N$ nodes into $K$ clusters (each with $c$ order nodes) to obtain the best score. The evaluation criteria of a cluster in BOBP is defined as the distance of picking route for retrieving its all items. It is noted that the essence of order is a set of items and the transformation from batch to item set is set union operation which is not differentiable. To tackle this specific problem, we firstly propose Balanced Task-oriented Graph Clustering Network (BTOGCN) which follows the Task-Oriented Prediction Network (TOPNet) \cite{chen2019task} with Graph Clustering \cite{wilder2019end}. The overview architecture of BTOGCN is illustrated in Figure \ref{fig:model}.  
BTOGCN learns a task-oriented estimator to directly estimate the downstream picking distance given graph embeddings, clustering results and labels, which guides the upstream clustering to achieve better performance in the downstream evaluation. In graph clustering, we use a differentiable k-means layer to cluster the order embeddings generated by a specifically designed type-aware heterogeneous graph convolutional network (HetGNN). We refer it as type-aware heterogeneous graph clustering networks. Moreover, we adopt a surrogate loss function to warm up the target estimator, making it sufficient-and-efficient to train the networks. The remainder of this section provides details on how the proposed method works. 


\subsection{Type-aware Heterogeneous Graph Clustering Networks On Balanced Order Batching Graph}

In this paper, we aim to learn the low-dimensional representation $h_v \in R^{n_{\phi(v)}}$ and apply it to the downstream node clustering task, where $n_{\phi(v)}$ is the dimension of embedding space for node type $\phi(v)$, for each vertex $v$ in the heterogeneous graph  let $N(v)$ be the set of adjacent vertexes of node $v$, $E(v)$ denotes the edges connected to $v$. Existing GCN-based works \cite{hamilton2017inductive,velivckovic2017graph,kipf2016semi} focus mainly on homogeneous graphs. Note that various relationship types can occur simultaneously in the $E(v)$, which brings challenges for heterogeneous graph embedding. To tackle these challenges, we propose a type-aware heterogeneous graph clustering method, namely HetGNN. Furthermore, for graph clustering task on heterogeneous graphs\cite{rozemberczki2019gemsec}, the graph is usually composed of millions of nodes and the objective is to detect community by training on subgraphs. However, in balanced order batching problem, the graph size is usually only a few thousand (order and item nodes) and it's not suitable to train on subgraphs since the task-based criteria is related to the whole graph. In particular, we use order embeddings as the input of clustering and we do cluster on each graph. 

\subsubsection{Type-aware heterogeneous graph convolutional network}
\begin{figure}[th]
	\centering
	\includegraphics[angle=0, width=0.90\columnwidth]{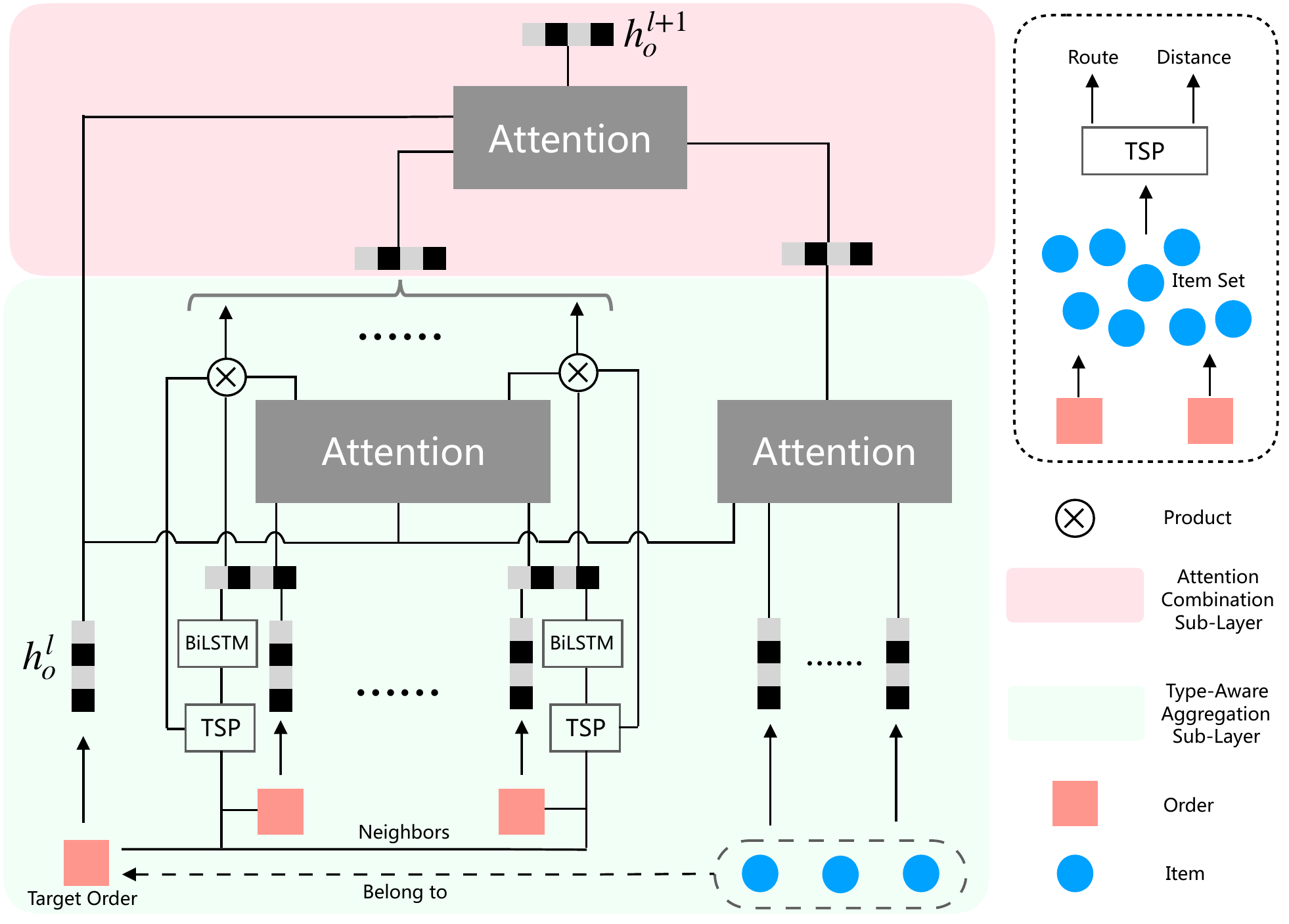}
	\caption{The data-flow of type-aware aggregation and attention combination sub-layer. It's an example of updating the embedding of an order node. 1) At type-aware aggregation sub-layer, attention mechanism is used to calculate the neighboring information according to linkage type. 2) At attention combination sub-layer attention mechanism is used to capture information among linkage types.}
	\label{fig:hetdata}
\end{figure}

GCN-based methods follow a layer-wise propagation manner, in which a propagation layer can be separated into two sub-layers: aggregation and combination. In the following section, we will concretely demonstrate how to 
customize type-aware aggregation sub-layer and attention combination sub-layer for a heterogeneous balanced order batching graph (HBOBG) with different types of node and edge attributes. The data-flow of type-aware aggregation and attention combination sub-layer is illustrated in Figure \ref{fig:hetdata}. 
\paragraph{\textbf{Type-aware Aggregation Sub-layer}}
Type-aware aggregation sub-layer is primarily motivated as an adaptation layer of Graph Neural Networks, which separately performs convolution operation on different types of graph neighborhoods. Aggregation operations vary by node and edge types. 

For an edge, only order-to-order ($oo$) edge has the hidden state, while other types of edges, i.e.,the order-to-item ($oi$) edge represent the inclusion relationship between order and items, and the item-to-item ($ii$) edge denotes the normalized picking distance between two items.
\begin{pequation}
	h_{e_r}^{l} = \left\{  
	\begin{array}{lr}
		h_{e_{oo}}^l,	\,\,if \, r=oo , & \\
		1\,\,\,\,\,\,\,\, ,	\,\,if \, r=oi, & \\
		d_{e_{ii}}\,\, ,	\,\,if \, r=ii. & 
	\end{array}
	\right. 
\end{pequation}
As a rule, the hidden states $h_{e_{oo}}^l$ is updated as the concatenation of previous hidden states of the edge itself and two nodes it links to. So the aggregation of edge hidden state is defined as:
\begin{pequation}
	h_{e_{oo}}^{l+1} = \sigma\left(W_{e_{oo}}^{l}.\left(h_{e_{oo}}^{l}||h_{o_1}^{l}||h_{o_2}^{l}\right)\right) \,,e_{oo}=(o_1,o_2,oo)
\end{pequation}
where $||$ denotes the concatenation operation.

For an order node $o \in \mathcal{O}$ and item node $i \in \mathcal{I}$, besides the information from neighbor nodes, the attributes of edges connected to them are also collected. The aggregated neighboring embedding $h_{N(o)}^l$ and $h_{N(i)}^l$ are calculated as

\begin{pequation}
	h_{N(o)}^{l+1} = \{h_{N(o),r}^{l+1}, \, \forall r \in R\}\\
	h_{N(i)}^{l+1} = \{ h_{N(i),r}^{l+1}, \, \forall r \in R\}
\end{pequation}
where $h_{N(o),r}^{l+1}$ and $h_{N(i),r}^{l+1}$ are aggregated neighbor embedding with linkage type $r$ and the detailed calculation process will be shown later.

Consider a vertex $j \in \mathcal{V}$, each neighboring vertex $m$ to vertex $j$ with linkage type $r$:
\begin{pequation}
h_{\phi(j),m,r}^{l+1}=W_{r}^{l+1}\left(h_m^{l}||\hat{h}_{e_r}^{l}\right),\,\, \forall e=(m,j,r) \in \mathcal{E}_{j,r}
\end{pequation}
and
\begin{pequation}
\hat{h}_{e_r}^{l} = \left\{  
\begin{array}{lr}
  h_{e_{oo}}^l,	\,\,if \, r=oo , & \\
  \varnothing,	\,\, others. & 
\end{array}
\right. 
\end{pequation}
The $|R|$ kinds of relations maintain different parameters $W_{\phi(j),r}$. To preserve the semantic of different types of relationships between nodes, we
utilize $|R|$ attention scoring functions \cite{vaswani2017attention} to match different relation patterns, i.e., $\mathcal{F}^{l+1}=\{f_r^{l+1}|r\in R\}$. For a vertex $j$ and linkage type $r$, an attention coefficient is computed for each edge $e=(m,j,r) \in \mathcal{E}_{j,r}$ in the form as:
\begin{pequation}
		attn_{m,j,r}^{l+1} = \sigma(f_r^{l+1}(h_{j}^{l},h_{\phi(j),m,r}^{l+1}))
\end{pequation}
where $\sigma$ is an activation function implemented RELU\cite{nair2010rectified}. The attention coefficient $attn_{m,j,r}^{l+1}$ indicates the importance of edge e to the target vertex j with linkage type $r$ . For simplicity, we adopt the same form of attention mechanism for all relation types but different in the parameters. A natural form of the attention scoring function is the dot product $f_r^{l+1}: h_{key} \times \mathcal{H}_{val} \to h_{val}$, which maps a feature vector $h_{key}$ and the set of candidates' feature vectors $\mathcal{H}_{val}$ to an weighted sum of elements in $\mathcal{H}_{val}$.
\begin{pequation}
	f_r^{l+1}(j,m) = h_{\phi(j),m,r}^{l+1} * {h_{j}^{l}}^T
\end{pequation}
where $*$ denotes the dot product operation. Notably, the trainable attention parameter shared by the same edge type $r$.  
Thereby, the softmax is applied over the neighboring linkages with type $r$ of vertex $j$ for the normalization of the attention coefficient. Moreover, for order-to-order linkages, the attention coefficient should be scaled with the normalized picking distance $d_{e_{oo}}$ before normalization

\begin{pequation}
	\alpha_{m,j,r}^{l+1} = exp(\hat{attn}_{m,j,r}^{l+1})/\sum_{e=(t,j,r)\in\mathcal{E}_{j,r}}exp(\hat{attn_{t,j,r}}^{l+1})\\
\end{pequation}
where
\begin{pequation}
	\hat{attn}_{m,j,r}^{l+1} = \left\{  
	\begin{array}{lr}
		attn_{m,j,r}^{l+1} * d_{e_{oo}},	\,\,if \, r=oo , & \\
		attn_{m,j,r}^{l+1},	\,\, others. & 
	\end{array}
	\right. 
\end{pequation}
Now we have the hidden states of neighboring nodes in the same low-dimensional space of the target node $j$, and weights of linkage $r$ associated with vertex $j$. Then with the in-degree distribution of vertex, the neighborhood aggregation for order and item node with linkage type $r$ can be performed as 

\begin{pequation}
	h_{N(o),r}^{l+1} = \sigma(\sum_{e=(m,o,r)\in\mathcal{E}_{o,r}}\alpha_{m,o,r}^{l+1}h_{\phi(j),m,r}^{l+1})\\
	h_{N(i),r}^{l+1} = \sigma(\sum_{e=(m,i,r)\in\mathcal{E}_{i,r}}\alpha_{m,i,r}^{l+1}h_{\phi(j),m,r}^{l+1})
\end{pequation}
\paragraph{\textbf{Attention Combination Sub-layer}}
After aggregating the neighbor's type-aware information, we adopt an attention combination strategy for the order and item nodes as
\begin{pequation}
	h_{o}^{l+1} = ATTN_{O}^{l+1}(h_{o}^{l},\{H_{N(o)}^{l+1},h_{o}^{l}\})\\
    h_{i}^{l+1} = ATTN_{I}^{l+1}(h_{i}^{l},\{H_{N(i)}^{l+1},h_{i}^{l}\})
\end{pequation} 
ATTN here is a naive dot-product attention mechanism mentioned above and the parameters involved in ATTN is different from node types at different layer, and we will not introduce it here. In a word, ATTN is used to combine context information among edge types. 

The whole algorithm is described in Algorithm \ref{alg-gcn} in Appendix \ref{app:algo}. Note that this method can actually be generalized to a hierarchical attention mechanism based on the meta-path schemes \cite{zhu2019relation}.

\paragraph{\textbf{Type-aware Sampling Strategy}}
With the proposed type-aware aggregation sub-layer and attention combination sub-layer, a whole-graph-batch training strategy can be conducted because all the entities should to be updated in one iteration to calculate the target-based criteria. For HBOBG, orders are connected with each other, so are the items. Regardless of the edge between the item and the order, the total edges is up to $624250$ fo a graph with $500$ orders and $1000$ items. Due to the time consumption, such massive edges should be reduced by sampling strategy. More details can be seen in Appendix \ref{app:sample}.
%
\paragraph{\textbf{Incorporate Graph Networks with Bidirectional-LSTM Model}}
The order-to-order edges should be converted into embedding before being merged with order features. In our specific graph, the $oo$ edge is represented by the picking route between two orders. Bi-directional Long Short-Term Memory (BiLTSM) \cite{huang2015bidirectional}is a satisfactory sequential model that balances effectiveness and efficiency. Therefore, we employ the BiLTSM model to get route embedding and integrate it to our graph neural network model as a part of an end-to-end framework.
The output of BiLTSM is then used as the embedding of the picking route. In detail,
\begin{pequation}
	h_{e_{oo}}^0 =BiLTSM(p_0,p_1,p_2,\dots,p_n)  
\end{pequation}
where $p_j$ represents item embedding of $j$-th item in picking route, and $h_{e_{oo}}^0$ is the initial embedding of edge $e_{oo}$ described above.

\subsubsection{Balanced K-means Layer}
Now, we can directly cluster orders as we get the low-dimensional and dense embedding of orders learned by HetGNN, then the cluster result, i.e. $K$ batches, will be used to calculate the total picking distance by non-differentiable set union operation and TSP algorithm process which will be discussed in the next section. Instead of using a clustering method such as k-means,we use a differentiable k-means version in \cite{wilder2019end} to enable the soft k-means assignment matrix $\hat{y}\in R^{N\times K}$ to be derived with respect to the order embedding vector $x$, i.e.,$\frac{\partial r}{\partial x}$ can be obtained. Furthermore, to achieve the balanced clustering, we adopt a global size constraint \cite{zhang2019framework} to help each cluster to be approximately the same size.
\begin{pequation}
	\mathcal{L}_G = \sum_{k\in \{1,\dots,K\}}{(\sum_{j}^{N}\hat{y}_{jk}/N-\frac{1}{K})}^2
\end{pequation}
However, the $\hat{y}$ is probability matrix and need a assignment algorithm to get discrete assignment result $y^{'}$ corresponding to the $K$ batches. We simply implement a greedy assignment algorithm, which iteratively choose the highest probability under the cluster size constraint.

\subsection{"Warm-up" based Task-Oriented Estimator}
The balanced order batching problem consists of two components: order clustering and picker routing. Order clustering is responsible for grouping orders into batches. And the picker routing firstly transforms the batch into an item set, then plans the picking route by TSP process for all of the items in that set. As the transformation from batch to item set is the non-differentiable set-union operation, the task-based criteria cannot be directly integrated into the end-to-end gradient based training process to guide the clustering.  
To automatically integrate the real task-based loss into our end-to-end learning process, we propose a task-oriented estimator network T, which takes both the extracted input feature $E_2(x)$, the encoding of our clustering assignment matrix $\hat{y}$
and labels $y$, to approximate the task-based loss $\mathcal{L}_{t}$ given a graph $x$, $\mathcal{L}_{t}$ is the total picking distance of the clustering solution generate by $y^{'}$ and can measure the order clustering performance directly:
\begin{pequation}
	\mathcal{L}_{t}(y^{'}) = \sum_{k=1}^{K}d_k \\
	\mathcal{T}(E_{2}(x),\hat{y},y) = (T_{1},\dots,T_{K}) \in R^{N\times K}
\end{pequation}
where $d_k$ is the picking route distance of orders in cluster $k$, which is calculated by solving a TSP. Furthermore, the estimation error can be derived as a standard mean square error between the task-based loss $\mathcal{L}_{t}$
and estimated score $\mathcal{T}$:
\begin{pequation}
	\mathcal{L}_{e}(\mathcal{T}(E_{2}(x),\hat{y},y),\mathcal{L}_{t}(y^{'})) = \frac{1}{K} \sum_{k=1}^{K}(d_k - \sum_{j=1}^{N}T_{jk} \hat{y}_{jk})^2
\end{pequation}
\subsubsection{Task-Oriented Objective: Estimated Clustering Loss}
Firstly, we borrow the idea of existing works, which mainly focus on using a surrogate loss function $\mathcal{L}_{s}$ to guide the learning process, where practitioners can either choose standard machine learning loss functions or other differentiable task-specific surrogate loss functions \cite{bengio1997using,elmachtoub2017smart,wilder2019melding,perrault2019decision,ferber2019mipaal}. For unsupervised clustering task, the training sample is not known previously, whereas the supervised classification task have ground truth as labels. If we have a clustering solution, we can naturally make it as a ground truth. The better the performance of the solution, the higher the accuracy of training. Nevertheless, there is no optimal solution for NP hard balanced batch processing, so we use a well-designed heuristic algorithm to generate a good initial solution to guide the clustering to meet the downstream evaluation criteria and to the task-oriented goal in the preheating stage. Therefore, the surrogate loss is a cross-entropy one and  $\mathcal{L}_{s}$ can be defined as follows:
\begin{pequation}
	\mathcal{L}_{s}(E_{1}(x),y,\hat{y}) = 
	-\frac{1}{N}\sum_{j=1}^{N}\sum_{k=1}^{K}\hat{y}_{jk}\log{(y_{jk})}
\end{pequation}
Besides, as the training goes on, we adopt a dynamic ground truth to make more effective clustering to step over the local minimal. In order to effectively avoid non-optimal local minimal and steadily increase task performance throughout training, we will update the ground truth if we get a better clustering solution with 
a significant improvement according to a paired t-test ($\alpha=5\%$). 

Consequently, in order to learn a universal task-oriented estimator for clustering tasks, BTOGCN hybridizes the surrogate loss function $\mathcal{L}_{s}$ and the estimated task-based loss $\mathcal{T}$ as estimated clustering loss: 
\begin{pequation}
	\mathcal{L}_{c} = \gamma*\mathcal{L}_{s} + \beta*\mathcal{T}
\end{pequation} 
The hyper-parameter $\gamma$ and $\beta$ depend on the estimation target error $\mathcal{L}_{e}$, i.e., when the $\mathcal{L}_{e}$ is below thresh $\varepsilon$, let $\gamma=0, \beta = 1$, otherwise let $\gamma=1, \beta = 0$. 
This exchange setting bridges the supervision from both the labels and the task-based criteria. In fact, it enables BTOGCN to "warm up" both the clustering and the task-oriented estimator T using the designed surrogate loss at the early stage.

Intuitively, by utilizing the supervision from dynamic labels mentioned above to "warm up" a reasonable clustering with the surrogate loss function, the task-oriented estimator will not be used to guide clustering to get more reasonable clusters until it estimates the task-based loss well enough, vice versa. In sum, a well-learned task-oriented estimator would also improve the clustering, which collaboratively forms a virtuous circle for the learning of both the task-oriented estimator network and the clustering network.
\begin{algorithm} 
	\caption{Balanced Task-oriented Graph Clustering Network.}       
	\label{alg-model} 
	\begin{algorithmic}[1]
		\Require $x$ is balanced order batching graph sampled form the training set. $y$ is the label generated by a heuristic method. $\mathcal{C}$ is the clustering network, significance $\alpha$ is the threshold for paired t-test
		\State Calculate soft clustering assignment matrix $\hat{y} = \mathcal{C}(E_1(x))$
		\State Execute greedy search on $\hat{y}$ to obtain the clustering results $y^{'}$ 
		\State Evaluate  $r$ using picking route distance calculated by TSP algorithm and
		get a real task-based loss $\mathcal{L}_t(r)=\sum_{j=1}^{K}d_k$.
		\If{$\mathcal{L}_t(y) > \mathcal{L}_t(y^{'})$ and OneSidedPairedTTest$({y},y^{'})<\alpha$}
		\State update the heuristic label $y=y^{'}$
		\EndIf
		\State Encode the input graph $x$ into $E_2(x)$ via a HetGNN $E_2$
		\State obtain the estimated task-based score $\mathcal{T}(E_{2}(x),\hat{y},y)$
		\State let the estimated clustering loss $\mathcal{L}_c=\gamma*\mathcal{L}_s+\beta*\mathcal{T}$
		\State update the clustering $\mathcal{C}$ and the HetGNN $E_1$ by minimizing $\mathcal{L}_c$
		\State Update the task-oriented estimator $T$, the HetGNN $E_2$ by minimizing the MSE between the task-based loss and estimated score  $\mathcal{L}_{e}(\mathcal{T}(E_{2}(x),\hat{y},y),\mathcal{L}_{t}(y^{'}))$
	\end{algorithmic}
\end{algorithm}

\section{Experiments}
\label{sec:exp}
In this computational experiment, we test our method and compare it with other existing methods on a set of real logistic orders, which is from real-world order data in Cainiao WMS. 

\subsection{Implementation Details}
\subsubsection{Dataset Description and Learning Problem}
We collect a graph dataset on balanced order batching problem based on real-world logistic order data in Cainiao WMS. In particular, for convenience, each graph is always constructed by 500 online orders. The task is simplified as finding a best solution which averagely assigns these 500 orders to 20 batches(i.e., each with 25 orders). The solution is evaluated by the total walking distance needed for retrieving all items of these 20 batches. We take the problem of picking up all items in one batch as a TSP and solve it with or-tools \cite{ortools}. The unit of walking distance is meter.
\subsubsection{Methods of Implementation}
Specifically, to demonstrate the effectiveness of task-oriented objective, we compare BTOGCN with the model which only utilizes supervised heuristic labels to predict a solution in Section 4.2.1. We call it Supervised-BTOGCN. 
However, it is noticed that except for the heuristics for BOBP, learning based approaches are the so-called two-stage methods, which can only get node embeddings or the clusters with no balance guarantee. Therefore, in order to get balanced clusters, we adopt a balanced k-means algorithm (BKM) \cite{malinen2014balanced} to conquer it. After that, we will calculate the total walking distance of picking up all items in these batches as the task score by calling the or-tools to solve TSP for each batch. 

As an alternative method, we consider the following well-known methods as our baseline, which are capable of producing clusters (batches) on our problem sets: 
\begin{itemize}
	\item BKM: run BKM with the orders' original high-dimensional features which contains the items' warehouse coordinations.
	\item Heuristics: a state-of-the-art algorithm proposed by \cite{ho2008order}. 
	\item AE+BKM: run BKM with the order embeddings produced by auto-encoder based method \cite{hinton2006fast}. 
	\item DEC+Assign: run DEC \cite{xie2016unsupervised} to get a relatively balance soft assignment matrix, then apply greedy assignment algorithm for it to satisfy the strict cluster size constraint

	
\end{itemize}
Notedly, since the DEC-based and AE-based methods focus on the node embeddings in  which the entire dataset shares same cluster centers, they cannot be applied in multi-graphs with different cluster centers.
\subsubsection{Hyper-parameters}
We use the 2-Layer HetGNN with 128 hidden units mentioned in 4.1 as graph embedding extractors $E1$ and a differentiable k-means layer to cluster the order embeddings. The graph embedding extractors $E2$ is a 2-Layer HetGNN with the same structure as $E1$.
To encoder the order-to-order edges, a 2-layer BiLSTM with 128 hidden units is used to get the initial route embeddings. The task-oriented estimator $T$ is 4-layer fully-connected neural networks with hidden units 128, 128, 128, 1. We train our model with Adam optimizer \cite{kingma2014adam} by initial learning rate of $10^{-3}$ and decay by a factor of $0.96$. More detailed implementation can be found in our code which will be shared upon paper acceptance. The implementation details of other baseline methods can be seen in Appendix \ref{app:h}.
 
\subsection{Results on single graphs}
Table 1 presents the results of our purposed model compared to non-learned baselines and state-of-art two-stage deep learning techniques. As mentioned above, AE+BKM is a two-stage clustering method which first learns node embedding individually and then do clustering on it, while DEC is an unconstrained end-to-end clustering method, whose clustering results can meet the balance requirements with the help of assignment algorithm, the whole method is denoted as DEC+Assign. AE+BKM and DEC+Assign are then "off-line" evaluated by an optimization algorithm, namely TSP. In single graphs experiments, we totally use 20 graphs with online data of different date and train our model on each separately. 

Firstly, it is seen that our proposed methods Supervised-BOTGCN and BOTGCN remarkably outperform the BKM, AE+BKM and DEC+Assign, which means the two-stage models will suffer from the lack of the guidance from task goal. In particular, our methods even obtain better solutions compared to the well-designed heuristics which demonstrates the superiority of our model which can transform a complex heuristics to a simple balanced k-means with no-inferior solution. Secondly, although the best batch in heuristics solution is better, the overall performance of all batches is no better than our method. In other words, BTOGCN places emphasis on the global optimization rather than only one or two batches having good quality. Moreover, due to the contribution of task-oriented network, BOTGCN has a further improvement compared to Supervised-BOTGCN. Finally, another interesting note is that the end-to-end deep clustering method DEC+Assign has the worst performance which supports our assumption that deep clustering methods will find the latent "clustering-friendly" representation space, however, it may not also be "friendly" with the cluster-based task. In addition, compared to BKM, the AE+BKM gains a significant improvement owing to the representation learning technique.

\begin{table}
	\caption{Results On Single Graph Set. "Avg Batch Score", "Max Batch Score" and "Min Batch Score" represent the average of the mean/maximum/minimum picking route distance of each graph in single graph set, respectively.}
	\scriptsize
	\label{tab:commands}
	\begin{tabular*}{\linewidth}{@{\extracolsep{\fill}}cccc}
		\toprule
		Methods & Avg Batch Score & Max Batch Score & Min Batch Score\\
		\midrule
		\texttt{Heuristics} & 835.76 & 1480.14 & \textbf{187.42} \\
		\texttt{BKM} & 901.17 & 1798.44 & 218.50 \\
		\texttt{DEC+Assign} & 950.09 & 1711.99 & 648.50\\
		\texttt{AE+BKM} & 884.30 & 1590.06 & 268.99 \\
		\texttt{Supervised-BTOGCN} & 835.11 & \textbf{1420.18} & 202.13\\
		\texttt{BTOGCN} &  \textbf{831.19} & 1459.32 & 196.60\\
		\bottomrule
	\end{tabular*}
\end{table}

\subsection{Generalizing across graphs}
Then, we investigate the generalization of our model on multi-graphs, as the order pool ceaselessly changes with the influx and consume of orders in warehouse. Specifically, we hope our approach can work on multi-graphs drawn from some order pool distribution so that it can be applied for on-line inference to unseen graphs. Indeed, we fix 40 graphs for training , 2 validation, and 8 test. As shown in Table \ref{tab:multi}, the task-oriented method surpasses all other methods thanks to the benefit of task-oriented estimator. We can conclude that the learned model successfully generalizes to completely unseen graphs. Although the heuristic method will obtain relatively good results, it's time-consuming. The result with limited time is even worse than the vanilla BKM in the quasi-real-time production systems. Taking advantage of estimating the task-based loss using a task-oriented estimator network, BTOGCN boosts the distance reduction by $3.99$m per graph compared with the model Supervised-BTOGCN trained with supervised label. 
\begin{table}
	\caption{Results On Multi-Graph Set. "Avg Batch Score", "Max Batch Score" and "Min Batch Score" represent the average of the mean/maximum/minimum picking route distance of each graph in multi-graph set, respectively.}
	\scriptsize
	\label{tab:multi}
	\begin{tabular*}{\linewidth}{@{\extracolsep{\fill}}cccc}
		\toprule
		Methods & Avg Batch Score & Max Batch Score & Min Batch Score\\
		\midrule
		\texttt{Heuristics} & 748.54 & 1451.20 & 168.72 \\
		\texttt{BKM} & 798.87 &1700.00  & 178.26 \\
		\texttt{Supervised-BTOGCN} & 752.40 &\textbf{1390.19}  & 167.61\\
		\texttt{BTOGCN} & \textbf{748.40}  & 1406.44 & \textbf{157.57}\\
		\bottomrule
	\end{tabular*}
\end{table}

\subsection{Simulation in warehouse}
\begin{figure}[th]
	\centering
	\includegraphics[angle=0, width=0.80\columnwidth]{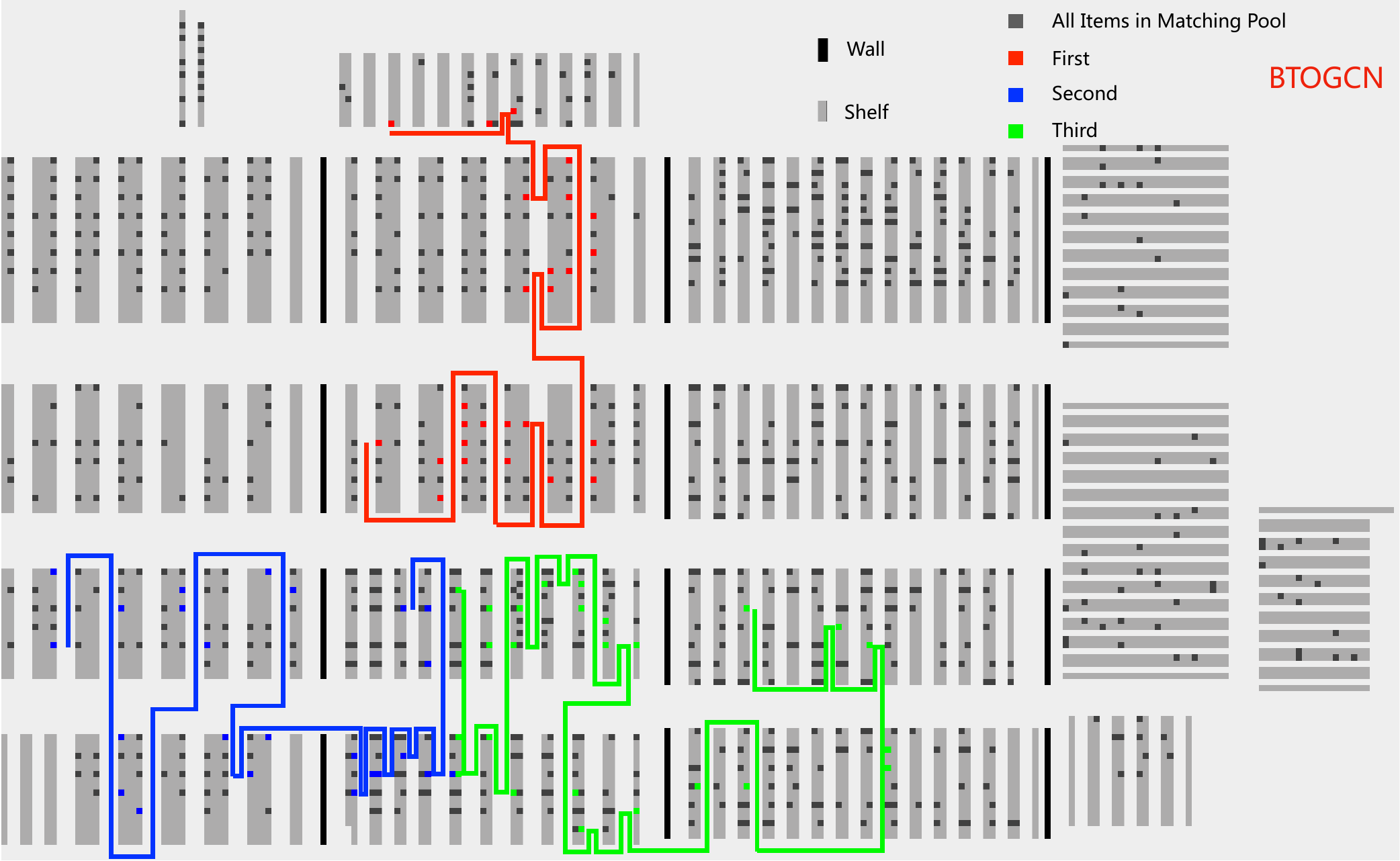}
	\caption{Top-3 pick lists generated by BTOGCN}
	\label{fig:result-example}
\end{figure}

The balanced order batching by different methods are executed in a warehouse by the order picker. We randomly choose a balance order batching graph and do simulation in a warehouse to indicate the higher performance of our purposed method. As shown in Figure \ref{fig:result-example}, for the sake of beauty, we just draw top-3 pick lists produced by BTOGCN on the map of the Cainiao warehouse in Hangzhou. Results of other methods can be seen in Appendix \ref{app:e}. Obviously, the results demonstrate that BTOGCN can produce more reasonable order batches than other methods. 

\section{Conclusion and Future Work}
\label{sec:conclusion}
This work aims to solve a practical problem existing widely in the warehouse named balance order batching problem. In this paper, we firstly propose a method combining machine learning with optimization algorithm to solve the balanced order batching problem (BOBP) which can directly and effectively improve the efficiency of warehouse operation. At the beginning, we give a formal problem definition, then reduce the BOBP to a balanced graph clustering optimization problem, whose task-based evaluation criteria is not differentiable. To tackle this specific problem, we propose a novel end-to-end approach, called Balanced Task-Oriented Graph Clustering Network (BTOGCN), which consists of an elaborately designed type-aware heterogeneous graph clustering networks and a task-based estimator, which can automatically integrate task-based evaluation criteria into the learning process with respect to the task-based goal. In our evaluation, we perform extensive analysis to demonstrate the highly positive effect of our proposed method targeting balance order batching problem. In our future work, we plan to use this framework to solve other combinatorial optimization problems, such as CVRP and bin packing problem.



\bibliographystyle{ACM-Reference-Format}
\bibliography{pickbill}

\clearpage
\appendix
\section{Algorithm.}
\subsection{Heterogeneous Graph Convolutional Networks}
\label{app:algo}
\begin{algorithm}                  
	\caption{Heterogeneous Graph Convolutional Networks on balanced order batching graph.}       
	\label{alg-gcn} 
	\begin{algorithmic}[1]
		\Require Balanced order batching graph  $\mathbb{G}(\mathcal{V}, \mathcal{E})$, $\mathcal{V}=\{\mathcal{O},\mathcal{I}\}$, where $\mathcal{O}$ is the set of candidate orders, $\mathcal{I}$ is the set of item nodes including in the candidate orders, edge types $R=\{oo,oi,ii\}$, number of layers $L$.
		\Ensure Hidden states of the $L$-th layer, include the hidden states of order nodes: $z_o , \forall o \in \mathcal{O}$ 
		
		\For {\emph{l = 1 to L}}
		\For {\emph{$e \in \mathcal{E}$}}
		\If{$r == oo$}
		\State $h_{e_{oo}}^{l+1} = \sigma(W_{e_{oo}}^{l}.f_e^{l+1}(h_{e_{oo}}^{l},h_{O(e_{oo})}^{l},h_{I(e_{oo})}^{l}))$
		\EndIf
		\EndFor
		\For {\emph{$j \in \mathcal{V}$}}
		\For {\emph{$\forall e=(m,j,r) \in \mathcal{E}_{j,r}$}}
		\If{$r == oo$}
		$\hat{h}_{e_r}^{l+1}= h_{e_{oo}}^{l}$
		\Else
		\State $\hat{h}_{e_r}^{l}=\varnothing$
		\EndIf
		\State $h_{\phi(j),m,r}^{l+1}=W_{r}^{l+1}(h_m^{l}||\hat{h}_{e_r}^{l})$
		\State $attn_{m,j,r}^{l+1} = \sigma(f_r^{l+1}(h_{j}^{l},h_{\phi(j),m,r}^{l+1}))$
		\If{$r==oo$}
		\State $attn_{m,j,r}^{l+1} = attn_{m,j,r}^{l+1} * d_{e_{mj}}$
		\EndIf
		\State $\alpha_{m,j,r}^{l+1} = softmax(attn_{m,j,r}^{l})$
		\EndFor
		\EndFor
		\For{\emph{$o \in \mathcal{O}$}}
		\State $h_{N(o),r}^{l+1} = \sigma(\sum_{e=(m,o,r)\in\mathcal{E}_{o,r}}\alpha_{m,o,r}^{l+1}h_{\phi(j),m,r}^{l+1})$
		\State $h_{o}^{l+1} = ATTN_{O}^{l+1}(h_{o}^{l},\{H_{N(o),oi}^{l+1},H_{N(o),oo}^{l+1},h_{o}^{l}\})$
		\EndFor
		\For{\emph{$i \in \mathcal{I}$}}
		\State $h_{N(i),r}^{l+1} = \sigma(\sum_{e=(m,i,r)\in\mathcal{E}_{i,r}}\alpha_{m,i,r}^{l+1}h_{\phi(j),m,r}^{l+1})$
		\State$h_{i}^{l+1} = ATTN_{I}^{l+1}(h_{i}^{l},\{H_{N(i),ii}^{l+1},H_{N(i),oi}^{l+1},h_{i}^{l}\})$
		\EndFor
		\EndFor
		\For {\emph{$o \in \mathcal{O}$}}
		\State $z_o = h_o^L$
		\EndFor
	\end{algorithmic}
\end{algorithm}
\subsection{Type-aware Sampling Strategy}
\label{app:sample}
We summarize the type-aware sampling strategy in the following:
\begin{itemize}
	\item For $oo$ and $ii$ edges, we choose the $M$-th closet orders/items. 
	\item For $oi$ edges, when the number of candidates is greater than the number of samples, i.e. $P$, we randomly sample $P$ items/orders. When the number of candidates is less than $P$, we pad them with placeholders, and ignore all the computations related to these placeholders.
\end{itemize} 

Different from random sampling strategy, we leverage the type information and propose a type-aware sampling. Our sampling strategy is more reasonable than random sampling in two aspects. First, different types of edges should have their own sampling strategies. For $oo$ and $ii$ edges, choosing closest orders/items is more reasonable than random sub-sampling since closest orders/items are more likely to cluster together. In the meanwhile, padding is more reasonable than re-sampling for $oi$ edges because the inclusion relation
between order and item are often sparse. Padding avoids changing neighborhood distribution compared to re-sampling. In this way, we achieve a comparable result with a small $M$ and $P$ thus saving training time as well as reducing memory consumption.
\section{B Experimental details}
\subsection{Hyper-parameters of baseline models}
\label{app:h}
We use majority of the publicly available code released by the author to report the performance of DEC. The auto-encoder we used in all experiments is the basic auto-encoder architecture, Its encoder is a fully connected multilayer perception with dimensions $m-1024-256-32$, where $m$ is the original data space dimension. The decoder is a mirrored version of the encoder.  All layers except the one preceding the output layer are applied a ReLU activation function. We train our model with Adam optimizer \cite{kingma2014adam} by initial learning rate of $10^{-3}$  with $\lambda = 0.001$, $\beta_1 = 0.9$, $\beta_2 = 0.999$. 
\subsection{Experimental Results}
\label{app:e}
Order picking simulation with other methods is shown as follows:
\begin{figure}[th]
	\centering
	\includegraphics[angle=0, width=0.80\columnwidth]{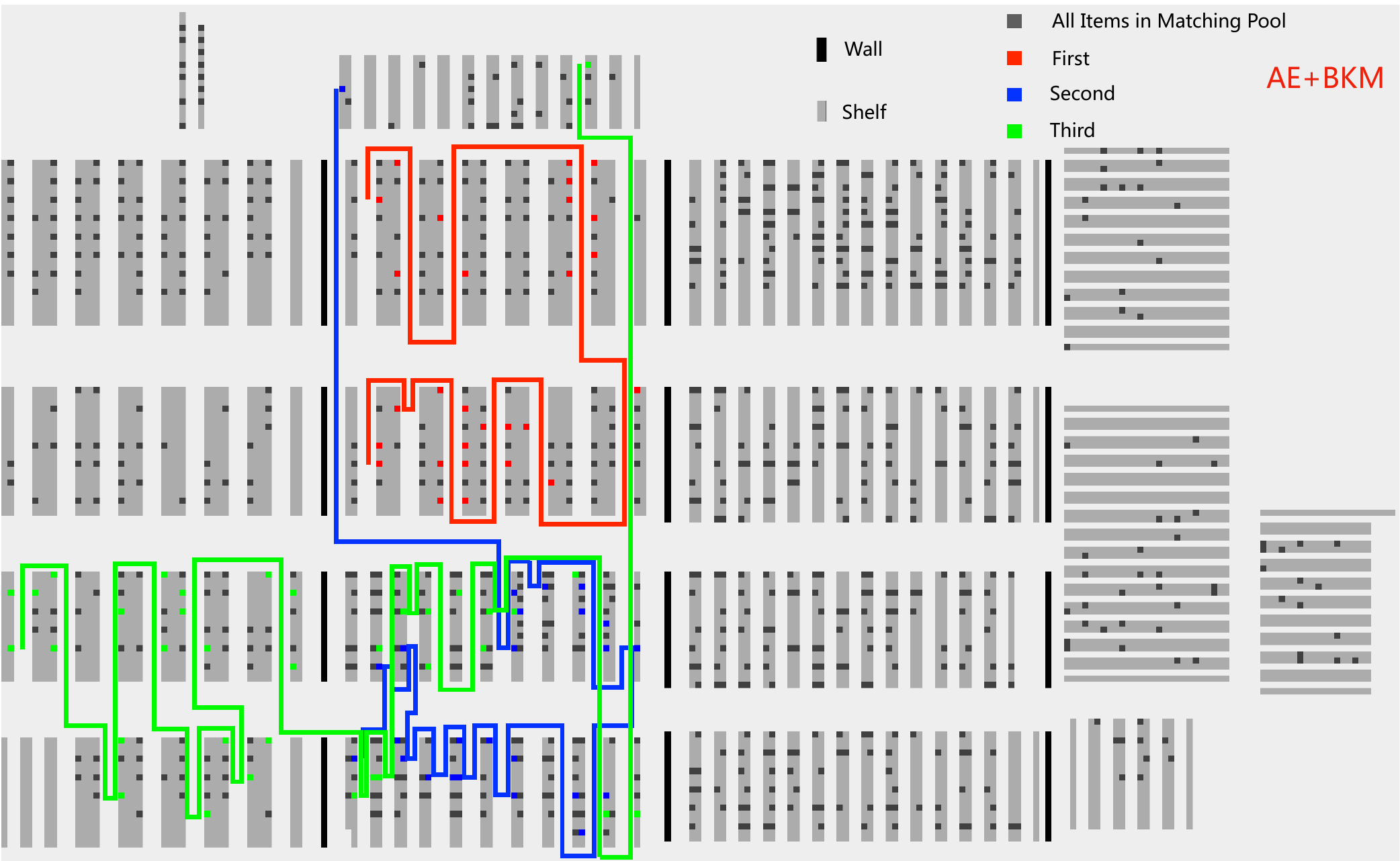}
	\caption{Top-3 pick lists generated by AE+BKM}
	\label{fig:result-example1}
\end{figure}
\begin{figure}[th]
	\centering
	\includegraphics[angle=0, width=0.80\columnwidth]{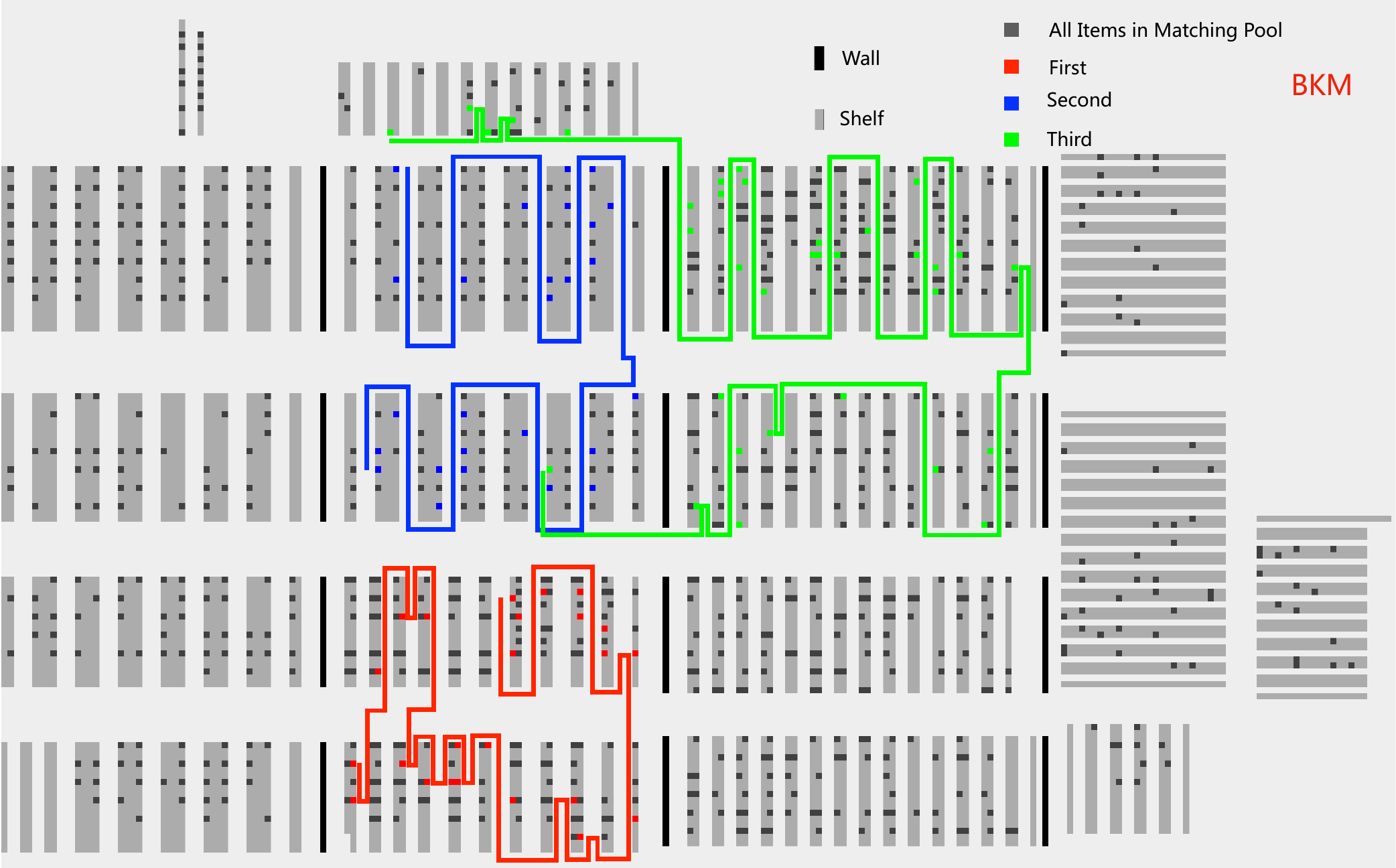}
	\caption{Top-3 pick lists generated by BKM}
	\label{fig:result-example2}
\end{figure}
\end{document}